\documentclass[journal]{IEEEtran}
\usepackage{amsmath}
\usepackage{url}

\usepackage{cite}

\ifCLASSINFOpdf
  \usepackage[pdftex]{graphicx}
\else
\fi
\begin{document}

\title{A Cycle-Resolved Cephalopod-Inspired Pulsed-Jet Robot with High-Volume Expulsion and Drag-Reduced Gliding}

\author{Yiyuan Zhang, \textit{Graduate Student Member, IEEE}, Anye Zhong, Junkai Chen, Wenci Xin \textit{Member, IEEE}

\thanks{This work was supported by the Ministry of Education, Singapore, through the REBOT project (“Rethinking underwater robot manipulation,” MOE-T2EP50221-0010); the Ministry of Foreign Affairs and International Cooperation, Italy, and the Agency for Science, Technology and Research, Singapore, through the DESTRO project (“Dextrous, strong yet soft robots,” R22I0IR124); the National University of Singapore Start-up Grant through the RoboLife project (“Soft robots with morphological adaptation and life-like abilities”); the National University of Singapore Bridging Fund through the project “AI-Driven Soft Robots for Marine and Unstructured Environments”; and the Singapore-MIT Alliance for Research and Technology (SMART) sub-award “AI-Driven Soft Robotics for Cleaning Tasks,” within the Mens, Manus, and Machina (M3S) project, funded by the National Research Foundation, Prime Minister’s Office, Singapore. \textit{(Yiyuan Zhang and Anye Zhong are co-first authors.) (Corresponding author: Yiyuan Zhang)}}

\thanks{Yiyuan Zhang, Anye Zhong, Junkai Chen are with the Department of Mechanical Engineering, College of Design and Engineering, National University of Singapore, Singapore. They are also with the Advanced Robotics Centre, National University of Singapore, Singapore. (e-mail: yiyuan.zhang@u.nus.edu)}

\thanks{Yiyuan Zhang and Wenci Xin are with the Singapore-MIT Alliance for Research and Technology Centre, Singapore.}
}

\markboth{Journal of \LaTeX\ Class Files,~Vol.~XX, No.~XX, XXX~XXXX}%
{Shell \MakeLowercase{\textit{et al.}}: Bare Demo of IEEEtran.cls for IEEE Journals}
\maketitle

\maketitle


\begin{abstract}
Cephalopod pulsed-jet locomotion is not a single isolated expulsion event, but involves jet expulsion, passive non-jetting motion, and mantle refilling. From this cycle-resolved perspective, this paper presents a cephalopod-inspired pulsed-jet robot with a origami mantle that enables large, actively driven, and geometry-guided body deformation. The proposed mantle integrates rigid folding panels with a compliant silicone framework, allowing a 75\% effective cavity-volume reduction during expulsion and reducing the projected cross-sectional drag area by approximately 75.7\% in the contracted gliding configuration. Using this platform, we formulate a cycle-resolved framework to separately investigate how expelled volume, glide duration, and refill pathway influence whole-cycle locomotion performance. Experiments show that the robot reaches a peak speed of approximately 0.5 m/s (3.8 BL/s) and an average speed exceeding 0.2 m/s (1.5 BL/s) within the first jetting cycle. The results further demonstrate the roles of high expelled-volume-ratio contraction in speed generation, reduced-drag-area gliding under different glide durations, and mantle-aperture-inspired passive inlet valves in assisting refill. This work provides both a robotic implementation of actively deformable cephalopod-like jet propulsion and a unified experimental platform for studying expulsion–gliding–refilling dynamics in pulsed-jet locomotion.
\end{abstract}

\begin{IEEEkeywords}
Bio-inspired robotics, soft robotics, underwater robots, pulsed-jet propulsion.
\end{IEEEkeywords}

%
\IEEEpeerreviewmaketitle

\section{Introduction}

\IEEEPARstart{C}{ephalopods} are among the most representative pulsed-jet swimmers in nature. Biological cephalopod jetting is not simply a uniformly repeated jetting event; when the locomotor cycle changes, the relative timing of jetting, refilling, and inter-pulse motion can also vary~\cite{Anderson2005}. Motivated by this phase-variable organization, we view one pulsed-jet cycle as an expulsion--gliding--refilling sequence and investigate how mantle deformation, gliding duration, and refill pathways jointly affect cycle-level locomotion performance. During expulsion, circular-muscle contraction reduces the mantle diameter and expels water through the funnel~\cite{kier2003muscle,trueman1968motor}; after expulsion, cephalopod can continue moving by inertia in a passive glide~\cite{gilly2012locomotion,flaspohler2019quantifying}; and during refilling, the mantle cavity re-expands mainly through elastic recoil, sometimes assisted by radial muscle contraction~\cite{kier2003muscle}. This cyclic view suggests that cephalopod-inspired pulsed-jet locomotion should be studied as a coordinated sequence rather than as an isolated jet-expulsion process.

In the expulsion phase, the mantle functions as an actively deformable chamber: the mantle contraction amplitude determines the amount of water expelled, while the resulting deformation also changes the swimmer's body morphology, as illustrated in Fig. ~\ref{fig:intro}(a). Fluid-dynamic studies on cephalopod-inspired pulsed-jet propulsion further suggest that propulsion performance is affected not only by the expelled jet, but also by body deformation during propulsion~\cite{RUIZ_WHITTLESEY_DABIRI_2011,steele2017added}. These observations motivate cephalopod pulsed-jet robots with mantles capable of large, controlled deformation, enabling substantial cavity-volume reduction for water expulsion and producing a contracted morphology for the subsequent locomotion cycle.

Several cephalopod-inspired robots have attempted to reproduce this expulsion mechanism using compliant bodies and soft actuation. A compliant mantle combined with cable-driven contraction and elastic recovery has been shown to generate effective underwater propulsion~\cite{6329967}. However, in fully soft cable-driven mantle designs, deformation can be strongly affected by material compliance and local cable loading. Serchi et al. showed that cable contraction can produce a lobed deformation of the shell cross-section and noted that robotic compression occurs near the cable attachment points, giving rise to buckling-like deformation of the external shell in those regions~\cite{serchi2013elastic}. As shown in Fig. ~\ref{fig:intro}(b), such localized deformation may leave undesired residual cavities during contraction, reducing the effective expelled volume that contributes to jet generation. To better guide deformation during large shape changes, later studies introduced structured compliant bodies and origami-inspired skins into pulsed-jet robot design~\cite{christianson2020cephalopod,yang2021origami}. These designs use ribs, endplates, or fold-pattern geometry to guide body deformation, rather than relying solely on homogeneous material compliance. Nevertheless, some existing robotic implementations differ from biological jetting in that their expulsion stroke depends strongly on stored elastic energy and passive release, whereas cephalopods actively contract the mantle musculature to expel water~\cite{kier2003muscle,trueman1968motor}. This difference motivates the development of a cephalopod-inspired pulsed-jet robot that combines active contraction with geometry-guided mantle deformation to support large-volume expulsion during the jetting phase.

In the gliding phase, the swimmer uses the velocity generated during expulsion to continue moving without active jetting. In biological locomotion, such intermittent motion has been interpreted as an energy-saving strategy~\cite{kramer2001behavioral,weihs1974energetic}, and similar ideas have also been explored in bio-inspired robotic swimming systems~\cite{liu2025bio,liu2026energy}. Although the fully contracted configuration appears at the end of expulsion, this morphology can be maintained throughout the following gliding interval. Therefore, a contracted body shape with a smaller effective frontal profile may reduce resistance during passive motion and extend useful gliding displacement. This raises a cycle-level question relevant to both robotic design and biological interpretation: how does glide duration trade off energy saving against average-speed loss when the post-expulsion body morphology is maintained during passive motion?

In the refilling phase, the mantle cavity resets for the next expulsion stroke. In squid, refill is enabled by mantle re-expansion through elastic recoil and, when needed, radial muscle activity~\cite{kier2003muscle}. The refill pathway is also functionally important: under a prescribed refill motion, a larger effective inlet can reduce inward-flow kinetic energy and internal dissipation, whereas a smaller outlet during jetting helps maintain jet momentum flux~\cite{bi2022role}. This suggests that cephalopod mantle apertures may create phase-dependent opening conditions between expulsion and refilling. In cephalopod pulsed-jet robots, passive one-way inlet valves may serve as simplified analogues of these apertures, providing a controllable platform to study how aperture-like refill pathways affect the overall pulsed-jet cycle.

\subsection{Our Contributions}

\begin{figure}[!htbp]
\centering
\includegraphics[width=\columnwidth]{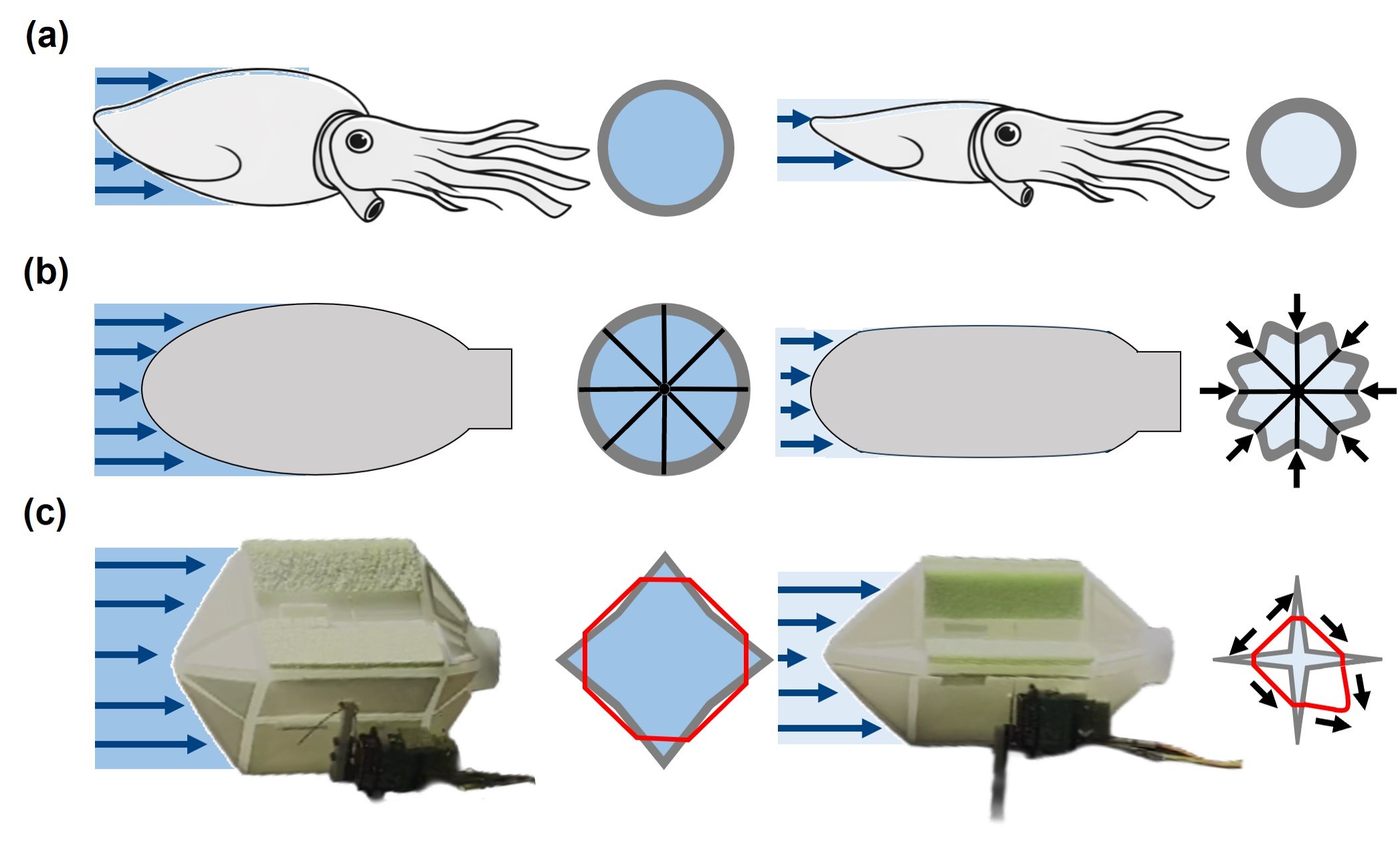}
\caption{From biological jet propulsion to robotic implementations. 
(a) Biological jet propulsion in cephalopods. 
(b) Previous pulsed-jet robot ~\cite{6329967,serchi2013elastic}. 
(c) Proposed pulsed-jet robot.}
\label{fig:intro}
\end{figure}

This paper presents a cephalopod-inspired pulsed-jet robot based on a sealed composite origami mantle, as illustrated in Fig.~\ref{fig:intro}(c). Rather than treating pulsed-jet locomotion only as a whole-cycle actuation problem, we decompose one robotic pulsed-jet cycle into expulsion, gliding, and refilling phases. This cycle-resolved framework allows the effects of volume expulsion, passive gliding, and aperture-assisted refill to be examined separately within a unified robotic platform.

The main contributions of this work are as follows:
\begin{enumerate}
    \item We propose a cycle-resolved framework for cephalopod-inspired pulsed-jet robot locomotion, in which one propulsion cycle is decomposed into expulsion, gliding, and refilling phases to evaluate their respective roles in whole-cycle performance.

    \item We develop a composite origami mantle that combines rigid folding panels with a compliant silicone framework, enabling large expelled volume ratio (EVR) and reduced frontal area during contraction. With this design, the robot achieves repetitive pulsed-jet propulsion, reaching a peak speed of approximately \(0.5~\mathrm{m/s}\) (\(3.8~\mathrm{BL/s}\)) and an average speed exceeding \(0.2~\mathrm{m/s}\) (\(1.5~\mathrm{BL/s}\)) within the first jetting cycle, while the cross-sectional area is reduced by \(75.7\%\) after expulsion.

    \item We establish the proposed robot as a unified platform for cycle-resolved study of cephalopod-inspired pulsed-jet locomotion. Using this platform, we examine three phase-level mechanisms: high-EVR expulsion for speed generation, the potential role of intermittent gliding in modulating the speed--energy trade-off between jetting events, and the possible contribution of aperture-assisted refill through mantle-aperture-inspired inlet valves.
\end{enumerate}

\section{Design and Fabrication}
\begin{figure}[!htbp]
\centering
\includegraphics[width=\columnwidth]{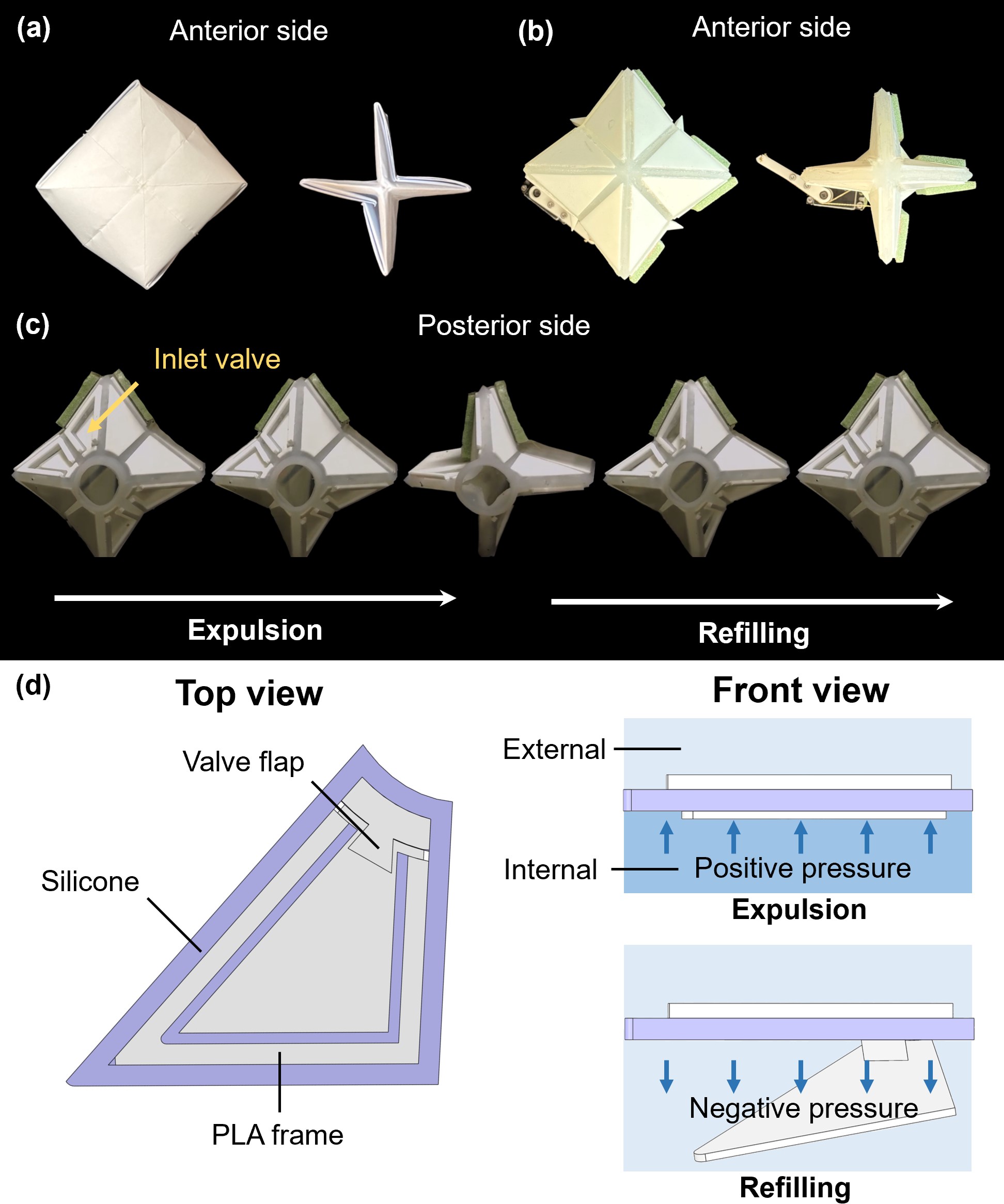}
\caption{Design and operating principle of the proposed pulsed-jet robot. (a) Expanded and contracted configurations of the anterior side of the waterbomb origami. (b) Expanded and contracted configurations of the posterior side of the proposed robot. (c) Opening--closing sequence of the posterior inlet valve during one actuation cycle. (d) Schematic of the inlet-valve working principle.}
\label{fig_design&manufacture1}
\end{figure}

The design of the robot was inspired by a common hand-folded origami pattern, namely the waterbomb origami configuration. As shown in Fig.~\ref{fig_design&manufacture1}(a), the left image shows the expanded configuration of the waterbomb origami, whereas the right image shows its contracted configuration, which forms a compact cross-shaped geometry, giving the contracted body a dart-like profile. To address the rigid-foldability constraints associated with the waterbomb pattern, a silicone framework layer is introduced to function as compliant joints. The silicone framework layer can both stretch and bend, thereby increasing the effective degrees of freedom, while also serving as a sealed flexible crease.
As shown in Fig.~\ref{fig_design&manufacture1}(b), the left image shows the robot in the expanded state, with a body length of 13.3~cm, and a cross-sectional area of 47.7~\(\mathrm{cm}^2\). The right image shows the robot in the contracted state and a cross-sectional area of 11.6~\(\mathrm{cm}^2\) (by manual measuring). Thus, the expanded-state cross-sectional area is approximately 4.1 times that of the contracted state. In this work, the body length (BL) used for body-length-normalized speed calculations is defined as the expanded-state body length.

To mimic the role of cephalopod mantle apertures as refilling pathways, passive one-way inlet valves were incorporated into the proposed robot. These valves are intended to remain closed during the expulsion phase to prevent fluid leakage through the inlet openings, while opening during the refilling phase to facilitate fluid intake into the mantle cavity, as shown in Fig.~\ref{fig_design&manufacture1}(c). During the expulsion phase, the positive pressure inside the mantle cavity presses the valve flaps against the silicone framework layer, thereby sealing the inlet openings and preventing fluid leakage. During the refilling phase, negative pressure is generated inside the mantle cavity, causing the valve flaps to open and allowing the surrounding fluid to be drawn into the cavity through both the inlet valves and the nozzle, as shown in Fig.~\ref{fig_design&manufacture1}(d).

\begin{figure}[!htbp]
\centering
\includegraphics[width=\columnwidth]{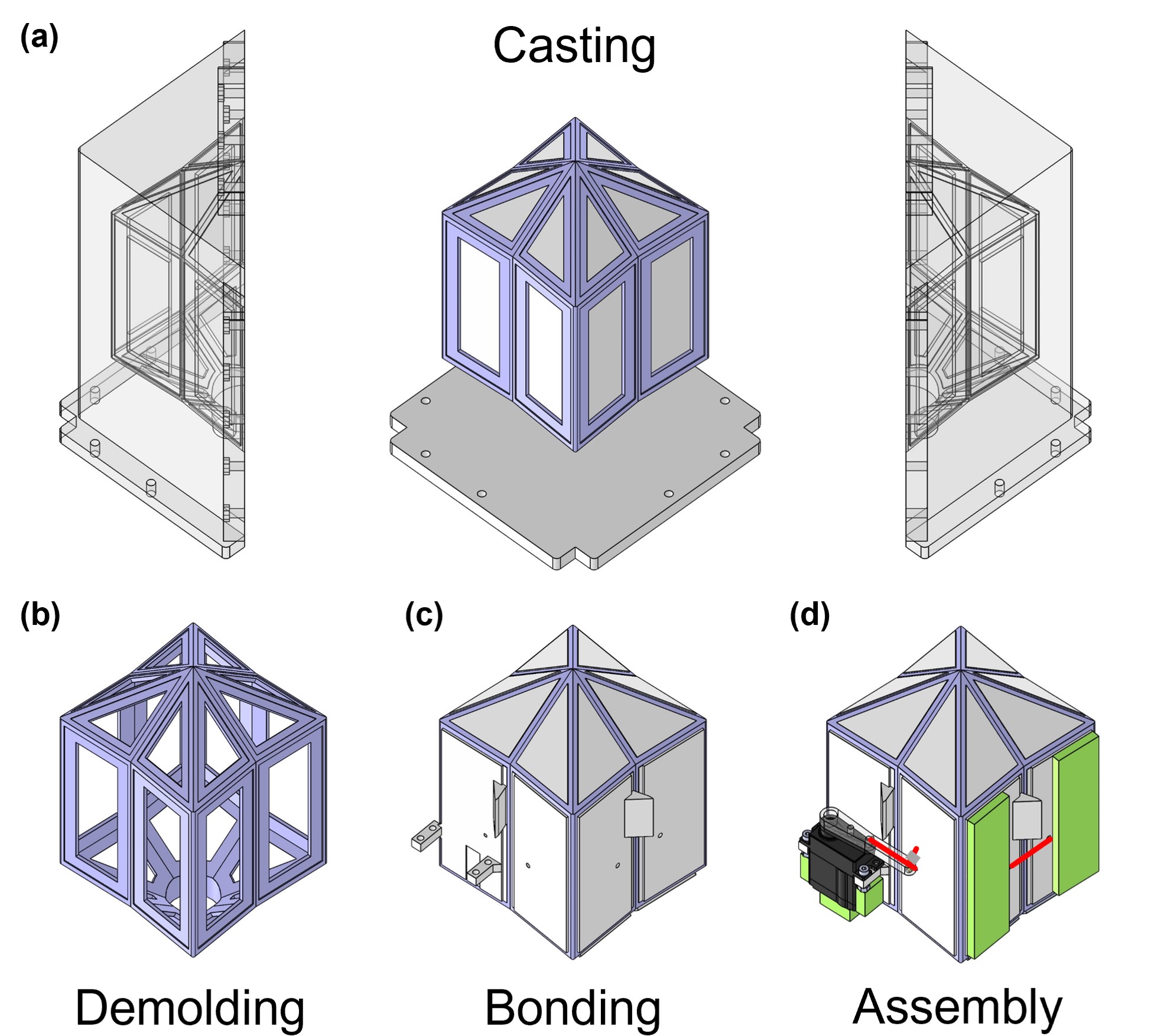}
\caption{Fabrication process of the origami-inspired robot. (a) Casting the silicone framework layer  using a 3D-printed mold. (b) Demolding the silicone framework layer . (c) Bonding the PLA panels onto the silicone framework layer . (d) Final assembly of the robot. }
\label{fig_design&manufacture2}
\end{figure}

Fig.~\ref{fig_design&manufacture2} illustrates the fabrication process of the robot. The mold for silicone framework layer casting consisted of three parts. Easy Release 200 release agent was first applied to each part. After the mold was assembled, mixed Dragon Skin 30 silicone was injected into the mold using a dispensing gun. The filled mold was then placed in a constant-temperature oven at 45 °C for 60 min before demolding, as shown in Fig.~\ref{fig_design&manufacture2}(a). After the silicone had cured, the connection between the sprue section and the mold base was manually broken. The mold component containing the sprue section was then removed through the larger opening in the silicone framework layer, as shown in Fig.~\ref{fig_design&manufacture2}(b). LOCTITE SF770 treatment was then applied to the reserved bonding regions of the silicone framework layer, and the 3D-printed PLA panels were bonded onto the silicone framework layer using LOCTITE 420 instant adhesive, as shown in Fig.~\ref{fig_design&manufacture2}(c). Finally, Kevlar cables (size no. 8, four-strand braid, diameter 0.45 mm) were threaded through the predesigned holes in the PLA panels as illustrates in Fig.~\ref{fig:intro}(c), a waterproof servo motor (PTK 9490 MG-D) was installed, and its arm was connected to the cables. Buoyant foam blocks ($0.08\,\mathrm{g\,cm^{-3}}$) were attached to selected panels and to the servo to ensure neutral buoyancy underwater and to provide a restoring buoyant moment that stabilized the robot with its upper side facing upward, as shown in Fig.~\ref{fig_design&manufacture2}(d). 

\section{Experimental Results}
\subsection{Experimental Comparison of Different Expelled Volume Ratios}
\begin{figure}[!htbp]
\centering
\includegraphics[width=\columnwidth]{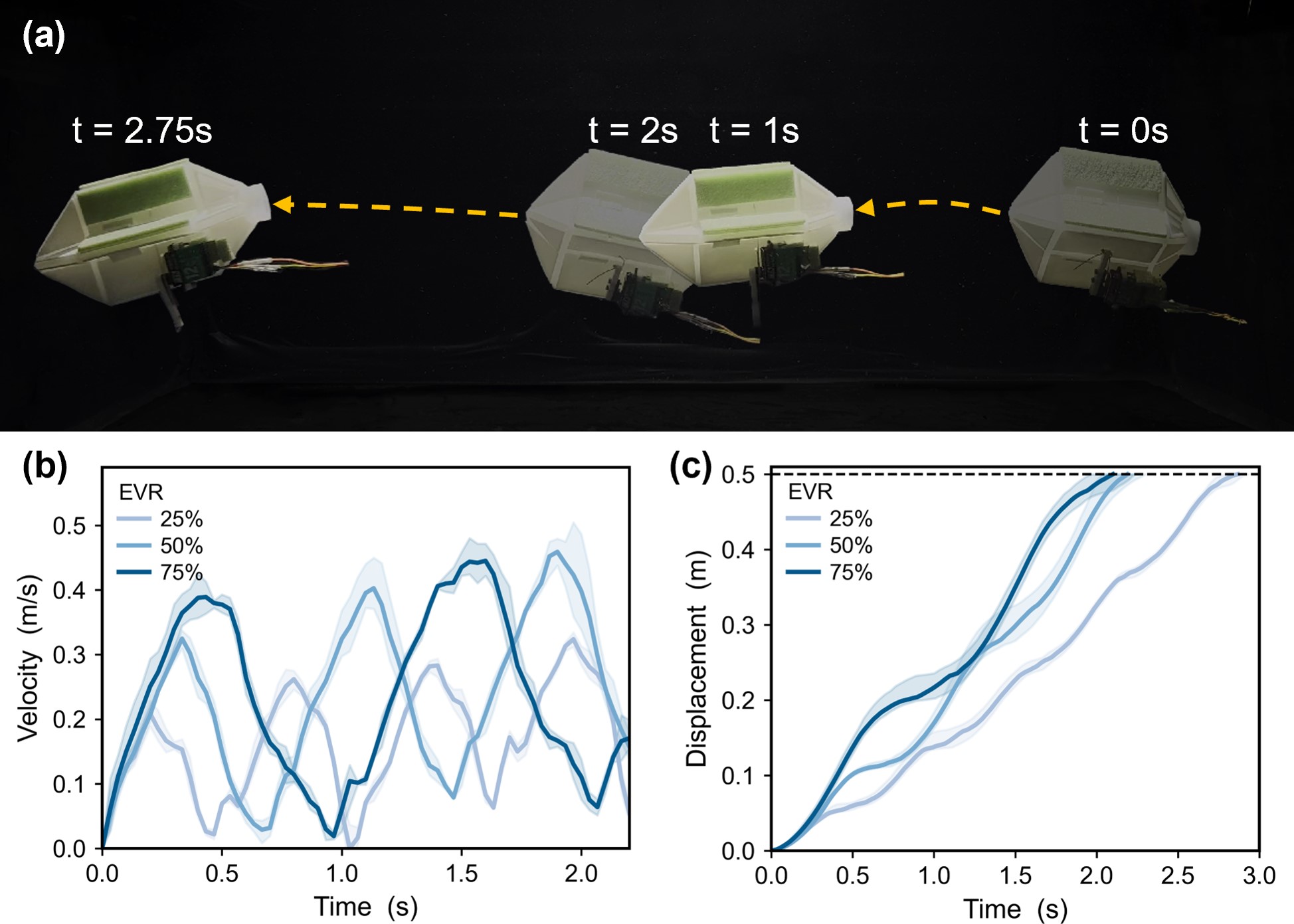}
\caption{Effect of EVR on swimming speed under the no-glide baseline condition. (a) Demonstration of the robot swimming in the experimental tank. (b) Velocity profiles of different EVRs over time. (c) Displacement profiles of different EVRs over time.}
\label{fig_exp1}
\end{figure}

Following previous pulsed-jet robot studies that commonly characterize propulsion through repeated expulsion--refill or inflation--deflation cycles~\cite{6329967,christianson2020cephalopod,giorgio2015thrust}, we first examined the effect of the robot's expelled volume ratio (EVR) under a no-glide baseline condition. The EVR is defined as
\begin{equation}
\mathrm{EVR} = \frac{V_{\mathrm{exp}}}{V_{\mathrm{tot}}}\times 100\%
\label{eq:evr}
\end{equation}
where \(V_{\mathrm{exp}}\) is the expelled fluid volume and \(V_{\mathrm{tot}}\) is the total fluid volume stored in the mantle cavity before expulsion. To evaluate the influence of EVR, a baseline cycle without an imposed gliding interval was first tested, in which the refill phase was initiated immediately after expulsion. The EVR was adjusted by setting the terminal position of the servo-driven contraction, with the nominal EVR values calibrated from expelled water volume and their repeatability reflected in the calibration and subsequent experimental error distributions. In all following experiments, the servo was controlled by PWM position commands without additional speed limitation, and the phase durations were determined from preliminary tests. Specifically, the robot position in the experimental water tank was tracked using Tracker Video Analysis software~\cite{tracker_osp}. Fig.~\ref{fig_exp1}(a) demonstrates the swimming performance of the robot in the tank, whose dimensions were 90 cm × 60 cm × 45 cm. As shown in Fig.~\ref{fig_exp1}(b), different EVRs led to different peak velocities. A larger EVR corresponded to a longer acceleration phase and a higher peak speed. During the refill phase, the magnitude of the velocity change also increased with EVR. During the first propulsion cycle, when the EVR was 25\%, the peak speed was 0.21 m/s; when the EVR was 50\%, the peak speed increased to 0.33 m/s; and when the EVR was 75\%, the peak speed further increased to 0.39 m/s. In addition, the velocity curve of each subsequent cycle was generally shifted upward relative to that of the preceding cycle. Fig.~\ref{fig_exp1}(c) further shows that the average speed also varied with EVR. When the displacement was fixed at 0.5 m, the groups with higher EVR exhibited higher average speed: the 25\% EVR case required 2.9 s and achieved an average speed of 0.17 m/s, the 50\% EVR case required 2.2 s and achieved an average speed of 0.23 m/s, and the 75\% EVR case required 2.1 s and achieved an average speed of 0.24 m/s.  A higher EVR indicates a larger expelled water volume under the same actuation-speed limit, thereby increasing the jet impulse. It also extends the later part of the expulsion process, during which the robot remains in a more contracted state with a smaller effective frontal area, further benefiting forward motion.
\subsection{Drag Measurement Experiment}
To assess how mantle folding affects frontal resistance, we conducted a terminal-fall drag characterization under different EVR configurations, to compare the drag-related geometric effect introduced by the deformation of the origami mantle structure. Since the actuator, buoyancy blocks, and cable arrangement remained nearly unchanged among different EVR cases, the variation in the identified drag parameter mainly reflects the geometric contribution of mantle folding.

The hydrodynamic drag was estimated using the form-drag relation:
\begin{equation}
F_D = \frac{1}{2}\rho C_D A U^2
\label{eq:drag}
\end{equation}
where \(F_D\) is the drag force, \(\rho\) is the fluid density, \(C_D\) is the drag coefficient, \(A\) is the projected frontal area, and \(U\) is the relative flow speed. In this work, \(C_DA\) is treated as an effective drag-area parameter for comparing different mantle configurations, as it provides a more meaningful hydrodynamic metric than projected area alone.

\begin{figure}[!htbp]
\centering
\includegraphics[width=0.8\columnwidth]{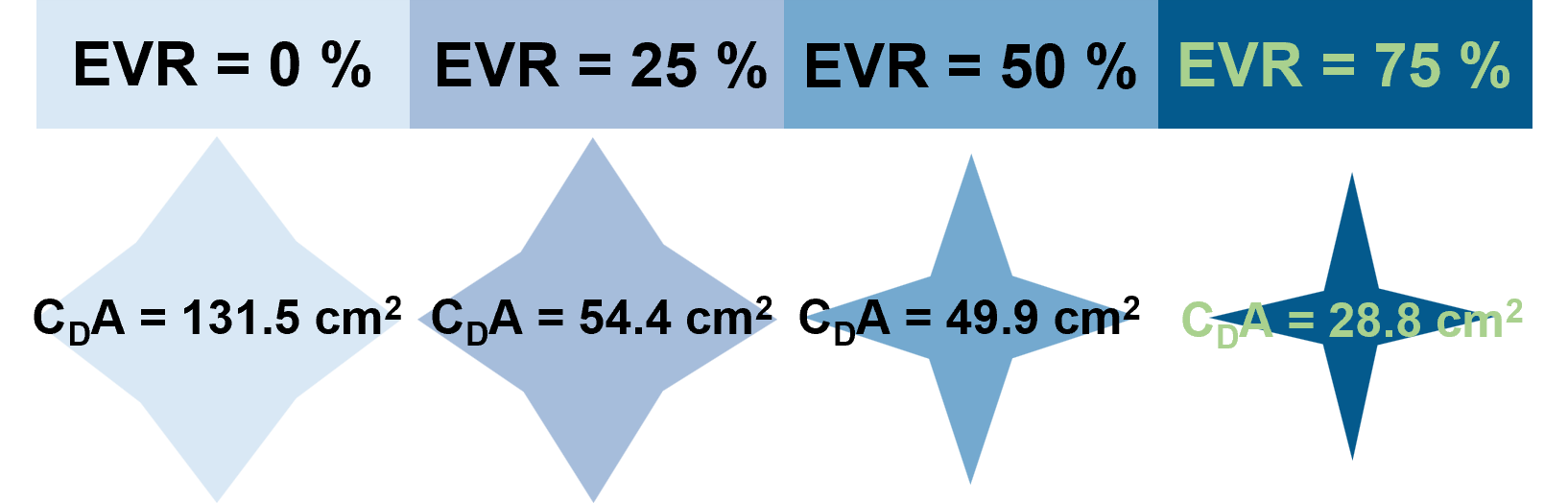}
\caption{Comparison of the effective drag-area parameter under different EVR configurations. 
Terminal-fall tests were conducted at EVR \(=0\%\), \(25\%\), \(50\%\), and \(75\%\).}
\label{fig_exp2}
\end{figure}

Inspired by the falling-body terminal-velocity method used for experimental drag-force estimation~\cite{OpenStax2016,yekutieli2005dynamic}, we estimated the effective drag-area parameter \(C_DA\) from the balance between form drag and the net downward force during vertical falling. Several buoyancy blocks were attached to the robot body so that the robot remained slightly negatively buoyant, while a restoring buoyant moment kept its central axis vertically downward. The magnitude of the net downward force was approximately \(0.01\,\mathrm{N}\). The robot was then released to fall vertically in the water tank. When the robot reached its terminal velocity \(U_t\), the form drag \(F_D\) balanced the net downward force \(F_{\mathrm{net}}\), defined as the difference between the robot weight \(W\) and the buoyancy force \(B\):
\begin{equation}
F_D = F_{\mathrm{net}} = W - B .
\label{eq:terminal_drag_balance}
\end{equation}
Since \(U_t\) could be measured using Tracker and the fluid density \(\rho\) was known, the corresponding effective drag-area parameter \(C_DA\) for each EVR configuration could be obtained by rearranging Eq.~\eqref{eq:drag}. The results for EVR values ranging from \(0\%\) to \(75\%\) are shown in Fig.~\ref{fig_exp2}(a)--(d), indicating that the identified \(C_DA\) decreased monotonically with increasing EVR.

Fig.~\ref{fig_exp2} further presents the ideal drag--velocity relationships for different EVR configurations, calculated from the identified \(C_DA\) values using Eq.~\eqref{eq:drag}. It can be seen that the EVR \(=75\%\) configuration has a substantially lower form drag than the other configurations at the same velocity. Combined with the results of the EVR swimming experiment, the EVR \(=75\%\) configuration not only achieves a higher velocity at the end of expulsion, but also exhibits a smaller effective drag-area parameter associated with the folded mantle geometry. This suggests that the highly contracted mantle configuration is more favorable for subsequent gliding.

\subsection{Gliding Strategy Experiment}
Based on the previous experiments, we identified a body configuration suitable for gliding. In this experiment, we investigated how glide time affects the locomotion of the pulsed-jet robot. Specifically, we compared the velocity--time curves, cost of transport (COT), and average speed under different glide-phase fractions (GPF) within a single jetting cycle, due to the limited size of the water tank. Two main questions were considered: We investigated whether increasing the gliding duration, corresponding to a larger GPF, reduces COT, and whether it necessarily lowers the average speed. The GPF is defined as follows:
\begin{equation}
\mathrm{GPF} = \frac{t_{\mathrm{gliding}}}{t_{\mathrm{expulsion}} + t_{\mathrm{gliding}} + t_{\mathrm{refilling}}}\times 100\%
\label{eq:gpf}
\end{equation}

\begin{figure}[!htbp]
\centering
\includegraphics[width=\columnwidth]{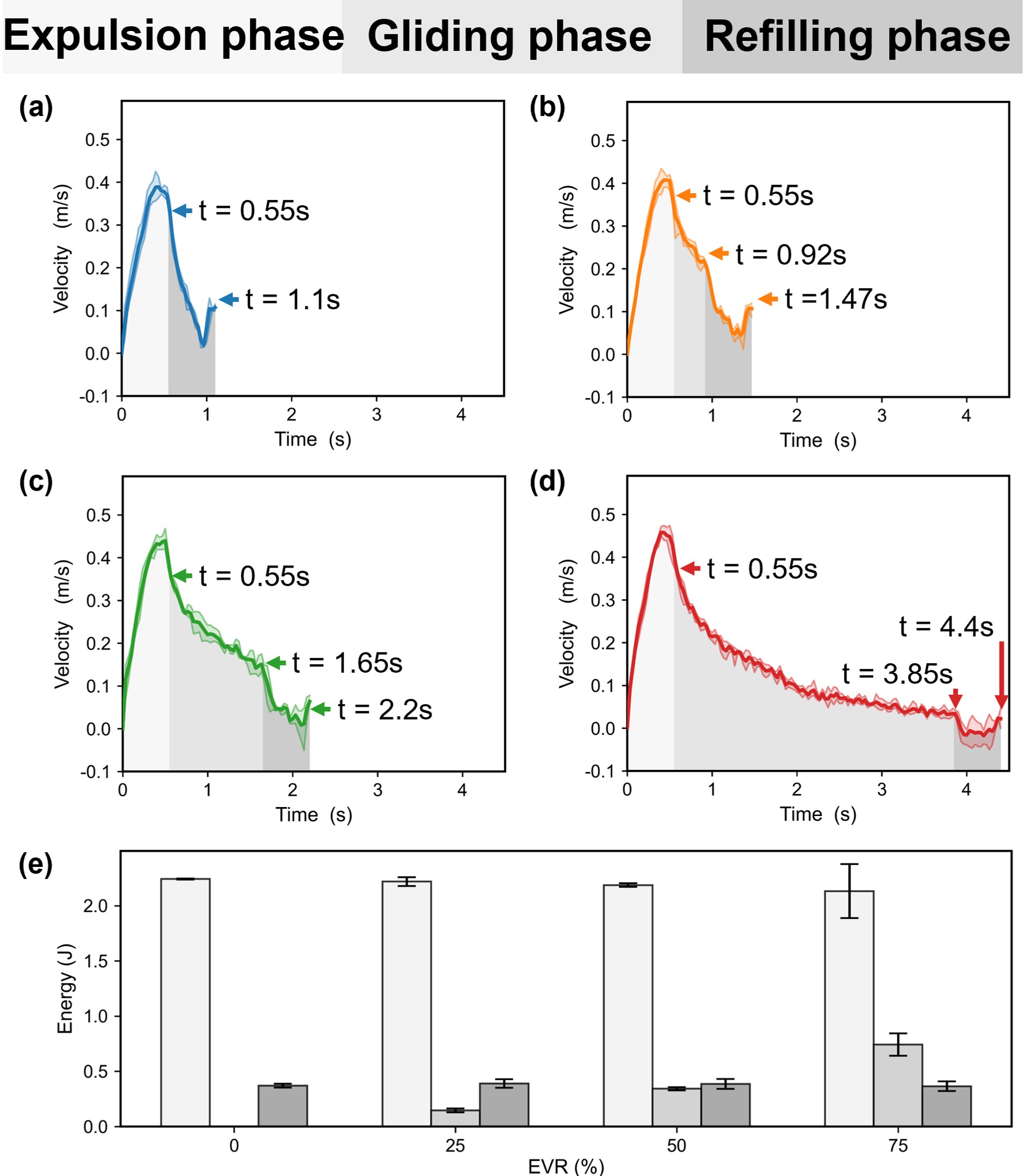}
\caption{Velocity--time curves and phase-wise energy consumption of the robot equipped with inlet valves under different gliding-phase fractions (GPFs). 
(a) GPF = 0\%, where the cycle consists of the expulsion phase followed directly by the refilling phase. 
(b) GPF = 25\%, with a short gliding phase inserted between expulsion and refilling. 
(c) GPF = 50\%, with an intermediate gliding duration. 
(d) GPF = 75\%, with an extended gliding duration. 
(e) Phase-wise energy consumption under different GPFs, comparing the energy consumed during the expulsion, gliding, and refilling phases.}
\label{fig_exp3}
\end{figure}

As shown in Fig.~\ref{fig_exp3}, the three shades of gray from light to dark represent the expulsion phase, gliding phase, and refilling phase, respectively. The durations of the expulsion and refilling phases were set to 0.55~s based on preliminary actuation tests, corresponding to a repeatable operating condition of the actuator. These two phase durations were kept constant in all cases so that the effect of the gliding-phase fraction (GPF) could be isolated. Fig.~\ref{fig_exp3}(a)--(d) correspond to GPF values of 0\%, 25\%, 50\%, and 75\%, respectively. It can be seen that the curves during the expulsion phase are approximately the same for different GPF. However, different GPF lead to different initial velocities at the beginning of the refilling phase: the larger the GPF, the lower the initial velocity in the subsequent refilling phase. This is because a larger GPF introduces a longer deceleration period before refill. The refill stage further shows that a higher initial velocity leads to a stronger deceleration effect. When the GPF is 0\% (no gliding), as shown in Fig.~\ref{fig_exp3}(a), most of the velocity achieved at the end of expulsion is lost during refill. In contrast, when the GPF is 75\%, as shown in Fig.~\ref{fig_exp3}(d), only a small portion of the remaining velocity is dissipated during refill. One reason is that a higher initial velocity produces larger drag, leading to more pronounced deceleration. Another important reason is the variable-mass effect: during refill, the robot must transfer part of its momentum to the incoming fluid in the mantle cavity so that the entrained water reaches the same velocity as the body. The faster the robot moves, the more momentum must be transferred to the incoming water, and the stronger the resulting deceleration. In addition, background flow has been shown to influence the expulsion process in jet-propelled swimming~\cite{luo2021jet}. By analogy, the background flow associated with different robot speeds may also affect refilling by altering the surrounding flow structure, pressure distribution over the body surface, and the resulting hydrodynamic force, thereby contributing to different deceleration behaviors.

As shown in Fig.~\ref{fig_exp4}(e), the COT decreases as the GPF increases within the tested range, but the reduction becomes much smaller when the GPF increases from 50\% to 75\%. This trend can be interpreted by considering the phase-wise energy consumption shown in Fig.~\ref{fig_exp3}(e). Within the tested GPF range, the expulsion phase consumes the largest amount of energy, and the values remain similar across different GPFs, approximately 2.2~J. The refilling-phase energy consumption is also comparable among different GPFs, approximately 0.4~J. In contrast, the energy consumed during the gliding phase increases with the gliding duration, with an approximately linear energy-consumption rate of 0.364~J/s.

Based on these observations, a simplified interpretation can be made by approximating the expulsion and refilling energy consumptions as nearly constant, while treating the gliding energy consumption as increasing approximately linearly with gliding duration because the servo must hold the contracted mantle configuration against elastic recovery. This loosely parallels biological systems, where maintaining an actively contracted muscular state can incur metabolic cost even in the absence of active propulsion. In addition, the displacement accumulated during the refilling phase is treated as a secondary contribution compared with the displacement accumulated during gliding. Under this approximation, the decrease in COT from GPF = 0\% to 50\% can be attributed to the relatively high robot speed during the gliding phase: the robot can travel a considerably longer distance within a short gliding duration, while only requiring a limited additional energy input. However, when the GPF increases to 75\%, the robot speed in the later part of the gliding phase becomes very low and approaches a nearly stopped state. As a result, further increasing the gliding duration provides only a small additional travel distance, while the gliding-phase energy consumption continues to increase. This explains why the COT reduction from GPF = 50\% to 75\% is much smaller. Therefore, the observed trend should not be interpreted as evidence that the COT monotonically approaches a limiting value as the GPF increases. Instead, it suggests a trade-off between the additional distance gained during gliding and the additional energy consumed during the extended gliding duration. Beyond the tested GPF range, because the gliding distance of a body in water is finite while the gliding-phase energy consumption continues to accumulate with time, the COT is expected to eventually reach a minimum and then increase. In the limiting case where the GPF approaches 100\%, the gliding duration would tend to infinity under the present definition, causing the energy consumption to diverge while the travel distance remains bounded; consequently, the COT would also tend to infinity.

As shown in Fig.~\ref{fig_exp4}(f), the average speed does not immediately decrease as the GPF increases; instead, it remains nearly unchanged at first and may even increase slightly. When the GPF is 50\%, the average speed (\(0.203~\mathrm{m/s}\)), although lower than the GPF is 0\% case (\(0.204~\mathrm{m/s}\)) and GPF is 25\% case (\(0.207~\mathrm{m/s}\)), still remains relatively high. However, when the GPF is 75\%, the average speed (\(0.129~\mathrm{m/s}\)) drops significantly. This behavior can be understood by noting that the robot undergoes two deceleration stages in each cycle: the gliding phase and the refilling phase. A longer gliding phase leads to greater deceleration during gliding, but reduces the severity of deceleration during refill. When the robot does not glide, the deceleration during refill is very strong and becomes the main factor limiting the average speed. In contrast, when the gliding duration is sufficiently long, the robot spends a substantial amount of time moving at a relatively low speed, which significantly reduces the average speed. These results suggest that there exists an intermediate GPF that balances these two effects and maximizes the average speed, rather than causing the average speed to decrease monotonically with increasing GPF. 

\subsection{Effects of Inlet Valves Experiment}
In this experiment, we investigated the effect of inlet valves, which serve as robotic analogues of cephalopod mantle apertures, on the robot's single-cycle locomotion. This experiment was built on the EVR characterization above; therefore, each single-cycle test was designed to begin and end with the mantle in the fully expanded state. Since the valve-free robot refilled more slowly through the jetting nozzle, a longer refilling duration was used to ensure complete elastic recovery of the mantle. This avoided incomplete refill, which would otherwise reduce the effective expelled volume and affect the subsequent jetting performance \cite{giorgio2015thrust}. Since the inlet valves mainly affect the refilling pathway, the expulsion and glide phases were kept consistent between the with-valve and valve-free configurations. Moreover, because the previous experiment showed that refill-induced deceleration is sensitive to the refill-onset velocity, each valve-free case was compared with the corresponding with-valve case under the same glide duration rather than the same GPF, thereby maintaining a comparable refill-onset velocity for the two configurations. Removing the inlet valves restricts the refill pathway; therefore, the refill command duration was increased from 0.55 s to 1.10 s to allow complete mantle recovery, thereby changing the total cycle duration. Therefore, COT and cycle-averaged speed were compared with respect to glide duration in order to ensure a fair comparison. .

\begin{figure}[!htbp]
\centering
\includegraphics[width=\columnwidth]{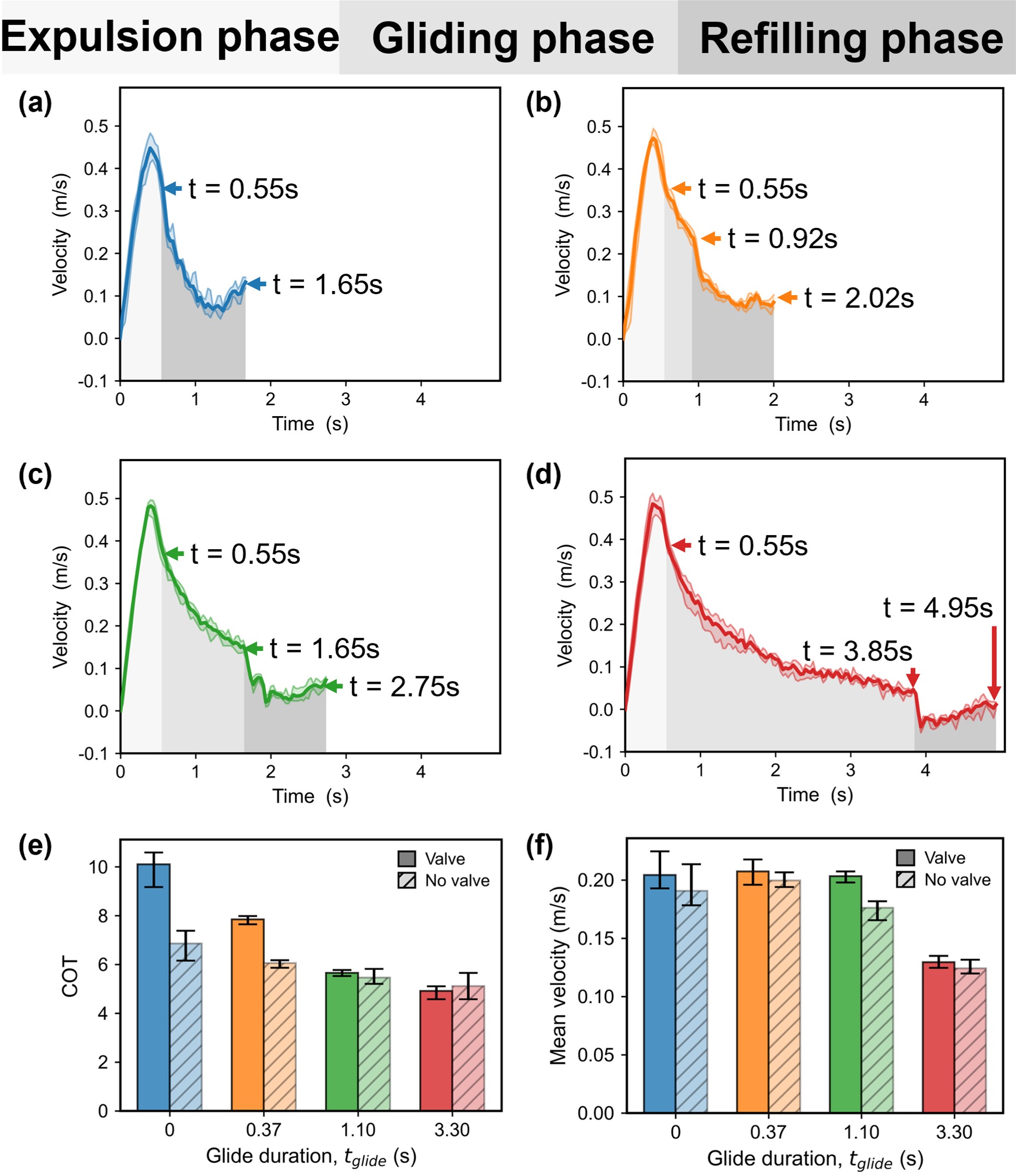}
\caption{Single-cycle swimming experiments of the valve-free robot and comparison with the robot equipped with inlet valves. The four glide durations correspond to the GPF \(=0\%\), \(25\%\), \(50\%\), and \(75\%\) conditions in the with-valve protocol. (a) Velocity--time curve for one jetting cycle at \(t_{\mathrm{glide}}=0~\mathrm{s}\). (b) Velocity--time curve for one jetting cycle at \(t_{\mathrm{glide}}=0.37~\mathrm{s}\). (c) Velocity--time curve for one jetting cycle at \(t_{\mathrm{glide}}=1.10~\mathrm{s}\). (d) Velocity--time curve for one jetting cycle at \(t_{\mathrm{glide}}=3.30~\mathrm{s}\). (e) COT versus glide duration. (f) Average speed versus glide duration.}
\label{fig_exp4}
\end{figure}

As shown in Fig.~\ref{fig_exp4}(a)--(c), for glide durations of \(0\), \(0.37\), and \(1.10~\mathrm{s}\), the most significant difference between the robot with inlet valves and the valve-free robot occurs during the refilling phase. The valve-free robot shows weaker refill-induced deceleration because water can only be drawn in through the nozzle. This restricted refill pathway reduces the amount of incoming water accelerated within the same time interval, thereby decreasing the associated momentum exchange and deceleration [Fig.~\ref{fig_exp3}(a)--(c)]. When the glide duration is \(3.30~\mathrm{s}\), both configurations enter refill at a velocity close to zero. In this case, the valve-free robot [Fig.~\ref{fig_exp4}(d)] shows slightly stronger deceleration than the robot with inlet valves [Fig.~\ref{fig_exp3}(d)]. A possible explanation is that, at such a low refill-onset speed, the restricted nozzle-only refill pathway prolongs the suction-induced deceleration, making its influence on the velocity curve more pronounced.
As shown in Fig.~\ref{fig_exp4}(e), when the glide duration ranges from \(0\) to \(1.10~\mathrm{s}\), the valve-free robot exhibits a lower COT than the robot with inlet valves. In this range, the valve-free robot still maintains a relatively high velocity during refill, and its longer refill duration leads to a larger single-cycle displacement. Although this longer refill process also introduces additional energy consumption due to servo holding, the displacement gain dominates the energy penalty, resulting in a lower COT. As shown in Fig.~\ref{fig_exp4}(e), when the glide duration reaches \(3.30~\mathrm{s}\), the COT relationship between the with-valve and valve-free robots is reversed. Therefore, the longer refill duration no longer produces a clear displacement gain, while the additional energy consumption remains. As a result, the robot with inlet valves shows a slightly lower COT at this glide duration. Nevertheless, both configurations still exhibit relatively low COT compared with the shorter-glide cases, indicating that the gliding strategy continues to provide an energy-saving benefit. This trend is consistent with the previous gliding-strategy analysis.

As shown in Fig.~\ref{fig_exp4}(f), the robot with inlet valves achieves a higher average speed than the valve-free version. This is because the valve-free version requires a longer refill duration due to ensure complete mantle recovery under the same EVR condition , and the additional time is spent at a speed significantly lower than the preceding cycle average, thereby reducing the average speed. 

By comparing the robots with and without inlet valves in Fig.~\ref{fig_exp4}(e) and Fig.~\ref{fig_exp4}(f), it can be seen that the configuration with inlet valves and a glide duration of \(1.10~\mathrm{s}\), corresponding to GPF \(=50\%\) in the with-valve protocol, not only achieves a near-minimum COT, but also maintains an average speed close to the maximum value. This is consistent with the original motivation for studying intermittent locomotion, namely, to substantially reduce energy consumption while causing only a small change in average speed. Therefore, based on the single-cycle experiments, we conclude that the robot with inlet valves and a glide duration of \(1.10~\mathrm{s}\) exhibits a clear overall advantage over the other tested cases.

\subsection{Multiple Cycle Jetting }
The previous single-cycle experiments were used to isolate the phase-level effects of expulsion, gliding, and refilling. Since periodic multi-cycle jetting can be viewed approximately as the accumulation of repeated single-cycle responses, these tests clarify the mechanisms underlying each cycle. However, repeated swimming may also be affected by accumulated velocity, tether interaction, and environmental disturbances. Therefore, selected multi-cycle experiments were conducted over a longer distance to validate whether the gliding strategy and inlet-valve effects observed in single-cycle tests remain effective during repeated jetting.

In this experiment, three representative cases were selected from the previous experiments to further validate the roles of the gliding strategy and inlet valves during multi-cycle swimming. For clarity, the robot with inlet valves and a glide duration of 0 s, the robot with inlet valves and a glide duration of 1.10 s, and the robot without inlet valves but with the same glide duration of 1.10 s are denoted as WV-0, WV-1.10, and NV-1.10, respectively. Their travel times over the same distance, as well as the corresponding energy consumption, were then compared.

\begin{figure}[!htbp]
\centering
\includegraphics[width=\columnwidth]{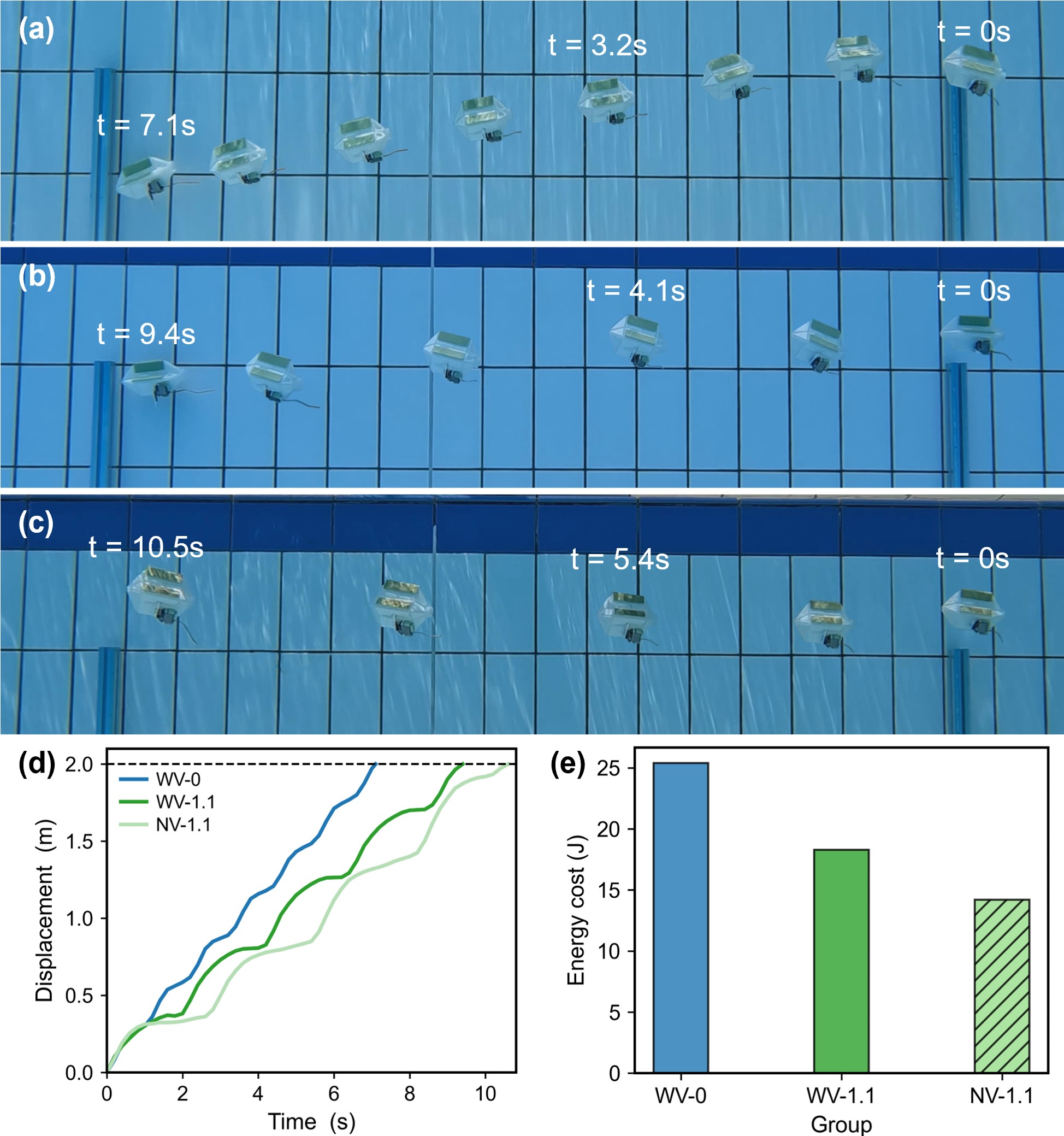}
\caption{Multiple-cycle swimming experiments validating the effects of the gliding strategy and inlet valves. WV-0, WV-1.10, and NV-1.10 denote the robot with inlet valves at a glide duration of \(0~\mathrm{s}\), the robot with inlet valves at a glide duration of \(1.10~\mathrm{s}\), and the robot without inlet valves at a glide duration of \(1.10~\mathrm{s}\), respectively. The glide duration of \(1.10~\mathrm{s}\) corresponds to the \(\mathrm{GPF}=50\%\) condition in the with-valve version. (a) Multi-exposure image of WV-0 swimming over \(2~\mathrm{m}\). (b) Multi-exposure image of WV-1.10 swimming over \(2~\mathrm{m}\). (c) Multi-exposure image of NV-1.10 swimming over \(2~\mathrm{m}\). (d) Comparison of displacement--time curves for the three cases. (e) Comparison of energy consumption for the three cases.}
\label{fig_exp5}
\end{figure}

 As shown in Fig.~\ref{fig_exp5}(a) and Fig.~\ref{fig_exp5}(b), WV-1.10 (COT = 7.65) exhibits a longer single-cycle gliding distance than WV-0 (COT = 10.63), while consuming less COT [Fig.~\ref{fig_exp5}(e)]. This result supports the effectiveness of the gliding strategy in reducing energy consumption within the tested range [Fig.~\ref{fig_exp3}(e)]. In addition, Fig.~\ref{fig_exp5}(a), Fig.~\ref{fig_exp5}(b), and Fig.~\ref{fig_exp5}(d) show that, over a longer travel distance, WV-0 (\(0.28~\mathrm{m/s}\)) is clearly faster than WV-1.10 (\(0.212~\mathrm{m/s}\)), whereas in the single-cycle experiments [Fig.~\ref{fig_exp3}(f)] its speed advantage was not significant. As shown in Fig.~\ref{fig_exp3}(a) and Fig.~\ref{fig_exp3}(c), WV-0 ends with a higher forward velocity (\(0.106~\mathrm{m/s}\)) than WV-1.10 (\(0.063~\mathrm{m/s}\)). Therefore, in repeated jetting, WV-0 enters the next cycle with a larger residual velocity, which accumulates over multiple cycles and results in a higher average speed over the 2~m distance.
As shown in Fig.~\ref{fig_exp5}(b) and Fig.~\ref{fig_exp5}(c), WV-1.10 reaches the destination faster than NV-1.10 (COT = 6.16 for NV-1.10) , but at the higher COT. This is consistent with the role of inlet valves identified in the previous experiments [Fig.~\ref{fig_exp3}(e)]. By shortening the refilling phase, the inlet valves increase the average speed. As discussed in the inlet-valve experiment range, at this refill-onset velocity, the longer refill process of the valve-free configuration can still provide additional forward displacement during refill, and this displacement gain can outweigh the extra energy consumed while the servo holds its position after returning. This interpretation is consistent with the trends in Fig.~\ref{fig_exp3}(e) and Fig.~\ref{fig_exp3}(f): the valve-free configuration can show a lower energy cost, whereas the with-valve configuration mainly benefits average speed by completing refill more quickly. 

Overall, these results suggest that the gliding strategy can substantially reduce COT with little change in average speed, whereas the inlet valves mainly serve to increase the average speed at the cost of higher energy consumption. It should be noted that the pool experiments were conducted under less controlled conditions than the indoor tank tests, including outdoor wind disturbance, background flow induced by other swimmers, and possible tether-related interference because the cable had to be manually carried behind the robot. Despite these uncertainties, the robot was still able to perform stable multi-cycle swimming over a 2~m distance and maintain strong speed performance. This indicates that the proposed robot does not rely on highly idealized laboratory conditions and has promising robustness for out-of-lab operation.

\section{Conclusion}
This letter presented a cephalopod-inspired pulsed-jet robot based on a composite origami mantle. By integrating rigid PLA panels with easy-to-demold slicone manbrane frame, the proposed design maintains structural simplicity and reliable motion while enabling controllable deformation and a relatively large volume-change capability (up to 75).

Experiments showed that the robot reached a peak speed close to \(0.5~\mathrm{m/s}\) (\(3.8~\mathrm{BL/s}\)) and an average speed exceeding \(0.2~\mathrm{m/s}\) (\(1.5~\mathrm{BL/s}\)) already within the first jetting cycle. These values compare favorably with representative cephalopod-inspired pulsed-jet robots whose reported speeds were evaluated over repeated actuation cycles. The average speed is about \(70.5\%\) higher than the normalized average speed of \(0.88~\mathrm{BL/s}\) reported by Giorgio-Serchi et al.~\cite{giorgio2016underwater}, while the normalized peak speed is about \(4.0\)-fold of the \(0.94~\mathrm{BL/s}\) reported by Christianson et al.~\cite{christianson2020cephalopod}. Further results showed that a higher expelled volume ratio improves propulsion performance, glide time changes the trade-off between average speed and energy consumption, and mantle apertures mimicing inlet valves further affect overall locomotion by influencing the refill process. These findings suggest that the performance of cephalopod-inspired pulsed-jet robots is determined not only by the expulsion phase, but also by the coordination among expulsion, gliding, and refilling.

We hope that this work provides an experimental platform for systematically investigating the roles of expelled-volume ratio, glide, and mantle apertures, which may offer transformative value for potential biological studies ~\cite{li2026bioinspired} and future underwater robot designs ~\cite{sun2024underwater }. The current system still has several limitations, including tethered external power and the lack of turning capability. Future work will focus on integrating onboard power and improving maneuverability.

\end{document}